\def\year{2020}\relax

\documentclass[letterpaper]{article} 
\usepackage{aaai20}  
\usepackage{times}  
\usepackage{helvet} 
\usepackage{courier}  
\usepackage[hyphens]{url}  
\usepackage{graphicx} 
\def\UrlFont{\rm}  
\usepackage{graphicx}  
\usepackage[utf8]{inputenc} 
\usepackage[T1]{fontenc}    
\usepackage{hyperref}       
\usepackage{url}            
\usepackage{booktabs}       
\usepackage{amsfonts}       
\usepackage{nicefrac}       
\usepackage{microtype}      

\usepackage{multirow}
\usepackage{graphicx}
\usepackage{algorithm}
\usepackage{algorithmicx}
\usepackage{amsmath,amssymb} 
\usepackage{algpseudocode}

\usepackage{color}

\usepackage{wrapfig}
\usepackage{subfigure}
\usepackage{algcompatible,lipsum}
\usepackage{cases}
\title{LPRNet: Lightweight Deep Network by Low-rank Pointwise Residual Convolution}
\author{Bin Sun, Jun Li, Ming Shao, Yun Fu}

\begin{document}

\maketitle

\begin{abstract}
Deep learning has become popular in recent years primarily due to the powerful computing device such as GPUs. However, deploying these deep models to end-user devices, smart phones, or embedded systems with limited resources is challenging. To reduce the computation and memory costs, we propose a novel lightweight deep learning module by low-rank pointwise residual (LPR) convolution, called LPRNet. Essentially, LPR aims at using low-rank approximation in pointwise convolution to further reduce the module size, while keeping depthwise convolutions as the residual module to rectify the LPR module. This is critical when the low-rankness undermines the convolution process. We embody our design by replacing modules of identical input-output dimension in MobileNet and ShuffleNetv2. Experiments on visual recognition tasks including image classification and face alignment on popular benchmarks show that our LPRNet achieves competitive performance but with significant reduction of Flops and memory cost compared to the state-of-the-art deep models focusing on model compression.
\end{abstract}

\section{Introduction}
During past years, deep convolutional neural networks (DCNN) have been widely used in many areas of machine learning and computer vision, such as face synthesis \cite{dollar2010cascaded}, image classification \cite{resnet}, pose estimation \cite{poserealtime}, and many more. However, the model complexity of DCNN in terms of time and space makes it hard for direct applications on mobile and embedded devices \cite{mobilenets,Mobilenetv2,Zhang_2018_shufflenet,shufflenetv2}. Therefore, there is an urge to design dedicated DCNN modules to reduce the computational cost and storage size for further applications on end devices. Furthermore, to make full use of existing networks, a general and efficient module is desired to replace the standard convolution module without changing the architectures.

\begin{figure}[!t]
\setlength{\abovecaptionskip}{-0.1 cm}
\setlength{\belowcaptionskip}{-0.3cm}
\centering
\includegraphics[width=0.9 \linewidth]{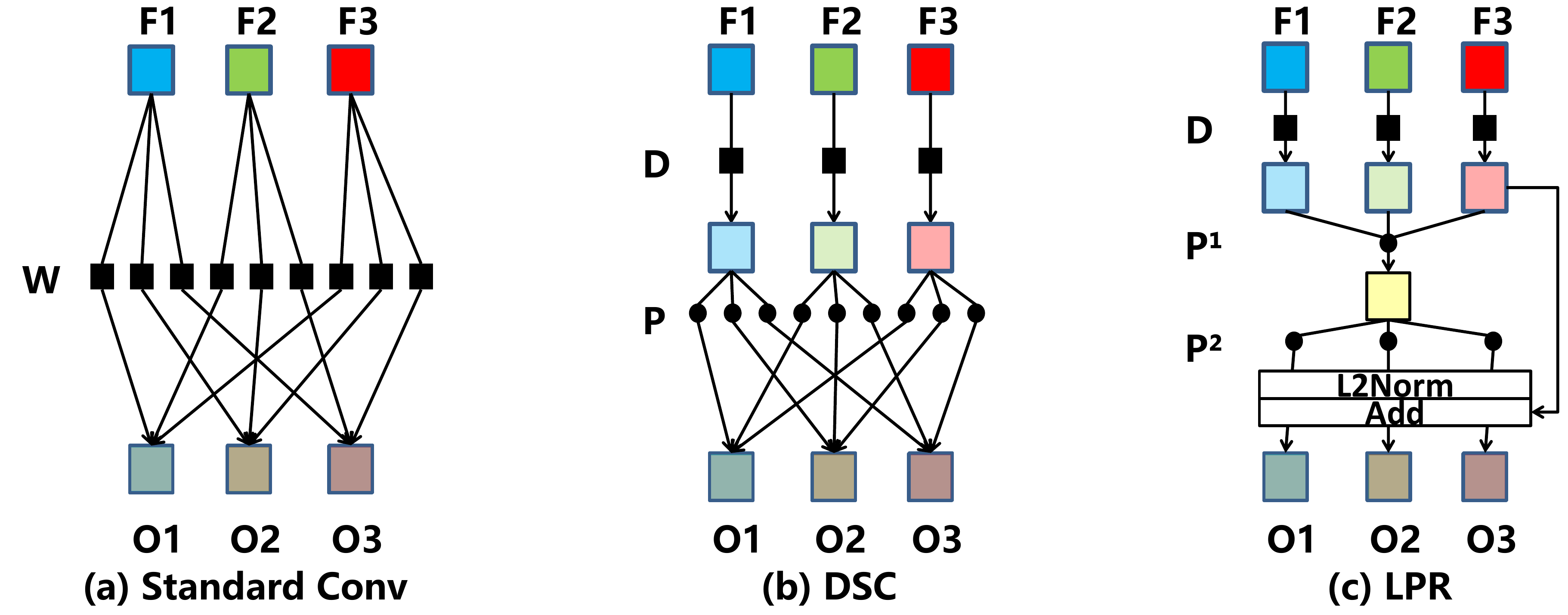}
\caption{
Illustration of light convolution architectures. (a): standard convolution. (b): depthwise separable convolution (DSC) module. (c): our low-rank pointwise residual (LPR) module.
}
\label{fig:basicmodule1}
\end{figure}

Standard convolution operation in Fig. \ref{fig:basicmodule1} (a) includes a large number of parameters (i.e., $nmk^2$, where $k\geq 3$ is the size of filters, $n$ and $m$ are the numbers of output and input feature channels or maps), which results in a high time and space complexity. To reduce the complexities, the standard convolution is divided into depthwise and pointwise convolutions, namely, depthwise separable convolution (DSC) \cite{sifre2014depthconv,mobilenets} in Fig. \ref{fig:basicmodule1} (b). Based on the DSC module, many lightweight networks are demonstrated, such as Xception model \cite{chollet2017xception}, SqueezeNet \cite{iandola2016squeezenet}, MobileNet \cite{mobilenets,Mobilenetv2}, ShuffleNet \cite{Zhang_2018_shufflenet,shufflenetv2}. In fact, the depthwise convolution applies a single filter to each input channel, and the pointwise convolution only uses $1 \times 1$ filters to compute the output features. Thus, the number of the parameters in the DSC module is significantly reduced to $mk^2+nm$. 
It should be noted that most of the popular DCNN models still use a large number of channels for better performance. Therefore, the pointwise convolution in DSC or relevant models still suffer from the higher time and space complexity. Besides, some modules like MobileNetv2 cannot fit into the existing networks without changing their architectures. Thus, a lighter model targeting at these problems is our research goal in this paper which may promote the deployment of more DCNNs to mobile applications.

To address the above problems, we introduce a novel DCNN parameters reduction module inspired by the principle of the low-rank CP-decomposition method \cite{speedingup-lowrank,CP-decomposition}. Instead of decomposing learned weight matrices, we apply the CP-decomposition on the layer design. Besides decomposing the conventional full-channel convolution into depthwise and pointwise convolutions, we will develop new learning paradigms for each of them, and thus reduce the overall model complexity. Following low-rank matrix decomposition idea, we are motivated to divide the large pointwise convolution into two small low-rank pointwise convolutions, as shown in Figure \ref{fig:basicmodule} (c). When its rank is $r<n,m$, the number of the parameters is further reduced to $mk^2+(n+m)r$. Furthermore, to compensate for the low-rankness of pointwise convolution and performance recession due to this compression, a residual operation through depthwise convolution is implemented to complement the feature maps without any additional parameters. To summarize, the contributions of this paper are:
\begin{itemize}
\item We propose a low-rank pointwise residual (LPR) module by automatically decomposing a larger pointwise convolution module into two low-rank matrices through the network learning, which significantly reduces the computational consumption. 
\item A residual learning mechanism is implemented to compensate for the information loss in pointwise convolution due to the matrix low-rankness, which guarantees the performance of our lightweight model without additional cost.
\item We embed our designed module in the network structure of MobileNet and ShuffleNetv2, and validate that our module can significantly reduce the parameters and FLOPs while keeping the performance with the same architecture. 

\item We also validate the correctness on image classification and face alignment tasks. On ImageNet dataset, while using much smaller parameters compared to the state-of-the-art, we achieved very competitive performance. On challenging face alignment benchmarks, our LPRNet obtains comparable results.
\end{itemize}

\section{Related Work}
In this section, we will first review the related work on the lightweight network construction. Then an overview of the state-of-the-art on image classification will be given. Last some related work on face alignment will be presented.
\subsection{Deep lightweight Structure}
In recent years, some methods have been emerged for speeding up the deep learning model. Faster activation function named rectified-linear activation function (ReLU) was proposed to accelerate the model \cite{relu}. Jin et.al. \cite{jin2014flattened} show the flattened CNN structure to accelerate the feedforward procedure.  In \cite{sifre2014depthconv} depthwise separable convolution was initially introduced and was used in Inception models \cite{ioffe2015batch}, Xception network \cite{chollet2017xception}, MobileNet \cite{mobilenets,Mobilenetv2}, and ShuffleNet \cite{Zhang_2018_shufflenet,shufflenetv2}, condensenet \cite{condensenet}.


Group Lasso \cite{grouplasso} is an efficient regularization for learning sparse structures. Jaderberg et.al. \cite{speedingup-lowrank} implemented the low-rank theory on the weights of filters with the separate convolution in different dimensions.  Liu et al. \cite{sparsecnn} implemented group Lasso to constrain the structure scale of LRA in CNN. To utilize DNN structure to different databases, Feng et al. \cite{feng2015learning} learned the appropriate number of filters in DNN. Wen et al. \cite{wensparsitylearning} applied group Lasso to regularize multiple DNN structures. In 2017, an architecture termed SVDNet \cite{svdnet} also considered matrix low-rankness in their framework to optimize the deep representation learning process. Different from SVDNet, our LPRNet utilizes the low-rank strategy as the guidance in the structure construction. 

\subsection{Image Classification}
Image classification has been extensively used to evaluate the performance of different deep learning models. For example, small-scale datasets \cite{cifar10} and large-scale datasets \cite{deng2009imagenet,imagenet} are often adopted as benchmarks in state-of-the-art works. In 2012, AlexNet was invented and considered as the first breakthrough DCNN model on ImageNet \cite{Alexnet}. Simonyan et al. later presented a deep network called VGG \cite{vgg} which further boosted the state-of-the-art performance on ImageNet. GoogLeNet \cite{googlenet} presented better results via an even deeper architecture. What followed is the widely adopted deep structure termed ResNet \cite{resnet} which enabled very deep networks and presented the state-of-the-art in 2016. Huang et al. further improved ResNet by densely using residuals in different layers. called DensenNet \cite{huang2017densenet} and improved the performance on ImageNet in 2017. Inception-v4 \cite{inceptionv4} is a novel structure that embraces the merits of both ResNet and GoogLeNet. While our LPRNet development is based on low-rank matrix decomposition, we also use residual term to compensate information loss due to compression. Most importantly, it retains the performance while reducing the parameters and computational burden.

\subsection{Face Alignment}
In conventional face alignment works, patch based regression methods were widely discussed \cite{ASM,CLM,AAM,baltruvsaitis2016openface} in past decades. In addition, tree-based methods \cite{kazemi2014one,LBF} with plain features attracted more attention, and achieved high speed alignment. Based on optimization theory, a cascade of weak regressors was implemented for face alignment \cite{xiong2013supervised}. Along with the rise of deep learning, Sun et al. \cite{sun2013deep} firstly utilized CNN model for face alignment with a face image as the input to CNN module, followed by regression on high-level features. It spawned considerable deep models \cite{trigeorgis2016mnemonic,liu2017dense,zhu2016face,Sun2018DeepE3,3DSTN,hyperface,kumar2018disentangling} that achieved good results on large pose face alignment. Besides, recently published large pose face alignment datasets with 3D warped faces for large poses \cite{zhu2016face}, or DNN structure Glass \cite{yang2017stacked,bulat2017far} have significantly promote the development and benchmarks in this field.
We also evalute our LPRNet on large pose face alignment problem to show its effectiveness and efficiency on the regression tasks, and mpressive performance on accuracy, speed, and size will demonstrated in experiment section.

\section{LPRNet}

In this section, we will elaborate the proposed LPRNet. First, we introduce the standard convolution \cite{vgg} and depthwise separable convolution from a matrix products perspective \cite{mobilenets}. Next, we propose our novel LPR structure and use it as building block in LPRNet. Finally, discussions and preliminary experiments of the LPRNet are shown. Notations of this paper have been summarized in in Table \ref{table:meaning}. 

\setlength{\tabcolsep}{1.2pt}
\begin{table}[!tp]
\setlength{\abovecaptionskip}{0.2cm}
\setlength{\belowcaptionskip}{-0.3 cm}
\begin{center}

\scalebox{0.9}{
\begin{tabular}{c|l}
\toprule[1.2pt]
Variable & ~~Description  \\
\hline
$S_{\text{F}}$&  size of one feature map \\
$S_{\text{k}}$&  size of one kernel  \\
 $C_{\text{in}}$ &  number of input channels \\
 $C_{\text{out}}$ &  number of output channels \\
 $D_{ij}$ &  $i$th weight for the $j$th feature in Depthwise layer \\
  $p_{ij}$ &  $i$th weight for the $j$th feature in Ponitwise layer\\
$W_{ij}$ &  $i$th weight for the $j$th feature in Standard Conv. \\
$\otimes$ & Kronecker Product \\
$F_{j}$ &  $j$th feature map of the input \\
\bottomrule[1.2pt]
\end{tabular}
}
\caption{Notations summary.}\label{table:meaning}
\end{center}
\end{table}
\subsection{Standard Convolutions (SConv)}
In traditional DCNNs, the convolution operation is applied between each filter and the input feature map. Essentially, the filter applies different weights to different features while doing convolution. Afterwards, all features convoluted by one filter will be added together to generate a new feature map. The whole procedure is equivalent to a series of matrix products, which can be formally written as:
\begin{equation}
\begin{bmatrix}
\begin{smallmatrix}
    W_{11}  & \dots  & W_{1m} \\
    \vdots  & \ddots & \vdots \\
    W_{n1}  & \dots  & W_{nm}
    \end{smallmatrix}
\end{bmatrix}\otimes \begin{bmatrix}
\begin{smallmatrix}
    F_{1} & F_{2} & \dots & F_{m}
    \end{smallmatrix}
\end{bmatrix}^{T},
\label{equ.3}
\end{equation}
where $W_{ij}$ is the weight of the $i$-th filter corresponding to the $j$-th feature map, $F_{j}$ is the input feature map, and $W_{ij} \otimes F_{j}$ means the feature map $F_{j}$ is convoluted by filter with the weight $W_{ij}$. In this paper, each $W_{ij}$ is a $3\times3$ matrix (filter), and all of them constitute a large matrix $[W_{ij}]$, or simply $W$.

\subsection{Depthwise Separable Convolutions (DSC)}

Depthwise Separable Convolution layers are the keys for many lightweight neural networks \cite{Zhang_2018_shufflenet,mobilenets,Mobilenetv2}. It has two layers: depthwise convolutional layer and pointwise convolutional layer \cite{mobilenets}.

\setlength{\tabcolsep}{1.0pt}
\begin{table*}[!tp]
\setlength{\abovecaptionskip}{0.1 cm}
\setlength{\belowcaptionskip}{-0.1cm}
\begin{center}

\scalebox{0.9}{
\begin{tabular}{c| c| c}
\toprule[1.2pt]
Modules &  computational cost (FLOPs) & Parameters  \\
\hline
SConv \cite{vgg} & $ S_{\text{k}} ^{2}\times S_{\text{F}} ^{2} \times C_{in} \times C_{out}$ & $ S_{\text{k}} ^{2} \times C_{in} \times C_{out}$ \\
\hline
DSC \cite{mobilenets}  & $ (\frac{1}{C_{out}} +\frac{1}{S_{\text{k}} ^{2}} ) \times  S_{\text{k}} ^{2}\times S_{\text{F}} ^{2}\times C_{in} \times C_{out}$& $ (\frac{1}{C_{out}} +\frac{1}{S_{\text{k}} ^{2}} )\times S_{\text{k}} ^{2}\times C_{in}\times C_{out}$\\
\hline
Mobilev2 \cite{Mobilenetv2}  & $ (\frac{e}{C_{out}} +\frac{e+1}{S_{\text{k}} ^{2}} ) \times  S_{\text{k}} ^{2}\times S_{\text{F}} ^{2}\times C_{in} \times C_{out}$& $ (\frac{e}{C_{out}} +\frac{e+1}{S_{\text{k}} ^{2}} )\times S_{\text{k}} ^{2}\times C_{in}\times C_{out}$\\
\hline
Shufflev2 \cite{shufflenetv2}  & $ \frac{1}{2}(\frac{1}{C_{out}} +\frac{1}{S_{\text{k}} ^{2}} ) \times  S_{\text{k}} ^{2}\times S_{\text{F}} ^{2}\times C_{in} \times C_{out}$& $ \frac{1}{2}(\frac{1}{C_{out}} +\frac{1}{S_{\text{k}} ^{2}} )\times S_{\text{k}} ^{2}\times C_{in}\times C_{out}$\\
\hline

LPR  & $ (\frac{1}{C_{out}}+\frac{2}{kS_{\text{k}} ^{2}} ) \times  S_{\text{k}} ^{2}\times S_{\text{F}} ^{2}\times C_{in} \times C_{out}$& $ (\frac{1}{C_{out}}+\frac{2}{kS_{\text{k}} ^{2}} )\times S_{\text{k}} ^{2}\times C_{in} \times C_{out}$\\
\hline
\multicolumn{3}{c}{$ S_{\text{F}} =14, S_{\text{k}} =3, C_{in}=256, C_{out}=256, k=8,e=4$}\\
\hline
SConv \cite{vgg} &$115,605,504$&$589,824$\\
\hline
DSC  \cite{mobilenets} &$13,296,640$&$67,840$\\
\hline
Mobilev2  \cite{Mobilenetv2} &$66,031,616$&$336,896$\\
\hline
Shufflev2  \cite{shufflenetv2} &$6,648,320$&$33,920$\\
\hline

LPR  &$3,662,848$&$18,688$\\
\bottomrule[1.2pt]
\end{tabular}
}
\caption{Comparisons on Flops and parameters. SConv, DSC, Shufflev2, Mobilev2, and LPR modules are used to build VGG, Mobilenetv1, Mobilenetv2, ShuffleNetv2 and our LPRNet respectively. }\label{table:computation compare}
\end{center}
\end{table*}

\textbf{Depthwise convolutional layer} applies a single convolutional filter to each input channel which will massively reduce the parameter and computational cost. Following the process of its convolution, we can describe the depthwise convolution using in the form of matrix products:
\begin{equation}
\begin{bmatrix}
\begin{smallmatrix}
    D_{11} & \dots  & 0 \\
    \vdots & \ddots & \vdots \\
    0 & \dots  & D_{mm}
    \end{smallmatrix}
\end{bmatrix}\otimes \begin{bmatrix}
\begin{smallmatrix}
    F_{1} & F_{2} & \dots & F_{m}
    \end{smallmatrix}
\end{bmatrix}^{T}
\label{equ.1}
\end{equation}
in which $D_{ij}$ is usually a $3 \times 3$ matrix, and $m$ is the number of the input feature maps.
We define $D$ as the matrix $[D_{jj}]$. Since $D$ is a diagonal matrix, the depthwise layer has much less parameters than a standard convolution layer.

\textbf{Pointwise convolutional layer} is using $1\times1$ convolution to build the new features through computing the linear combinations of all input channels. It follows the fashion of traditional convolution layer with the kernel size set to $1$. 
Following the process of its convolution, we can describe the pointwise convolution in the form of matrix products:
\begin{equation}
\begin{bmatrix}
\begin{smallmatrix}
    p_{11}  & \dots  & p_{1m} \\
    \vdots  & \ddots & \vdots \\
    p_{n1}  & \dots  & p_{nm}
    \end{smallmatrix}
\end{bmatrix}\otimes \begin{bmatrix}
\begin{smallmatrix}
    F_{1} & F_{2} & \dots & F_{m}
\end{smallmatrix}
\end{bmatrix}^{T}
\label{equ.2}
\end{equation}
in which $p_{ij}$ is a scalar, $m$ is the number of the input feature maps, and $n$ is the number of the output. The computational cost is $S_{\text{F}} \times S_{\text{F}} \times C_{in}\times C_{out}$, and the number of parameters is $C_{in}\times C_{out}$. We define $P \in \mathbb R^{m\times n}$ as the matrix $[p_{ij}]$. Since the depthwise searable convolution composed with depthwise convolution and pointwise convolution, it can be represented as:
\begin{equation}
F_m^{out} = W \otimes F^{\text{in}}_{m}\approx (P \times D )\otimes F^{\text{in}}_{m}
\label{equ.4}
\end{equation}



\subsection{LPR Structure}
In this subsection, we will detail the proposed LPR module, which has been shown in Figure \ref{fig:basicmodule} (c). Recall in the previous section, the depthwise convolution can be considered as the convolution between a diagonal matrix $dig(D_{11},...D_{mm})$ and a feature map matrix $[F_{1} ... F_{m}]$. We will keep this procedure but further explore the pointwise convolution in the following manner. To further reduce the size of matrix $P$, we will pursue a low-rank decomposition of $P$ such that $P \approx P^{(2)} \times P^{(1)}$, $P^{(1)} \in \mathbb R^{r\times m}$, $P^{(2)} \in \mathbb R^{m\times r}, r \ll m $. Clearly, the highest rank of this approximation is $r$, and the size of $m \times r$ is much smaller than $m^2$. Thus, we can naturally convert the original DSC module to:
\begin{equation}
F^{P}_{m}=(P^{(2)}_{mr}P^{(1)}_{rm})\otimes F^{D}_{m}, 
\label{equ.7}
\end{equation}
where $F^{P}$ means the output features after this new low-rank pointwise convolution operation. While using the strategy above may reduce the parameters and computational cost, it may undermine the original structure of $P$ when $r$ is inappropriately small, e.g., $r< \text{rank}(P)$. To address this issue, we propose to add a term $F_m^{Res} = D \otimes F_m^{in} $, i.e., the original feature map after the depthwise convolution with $D$. This ensures that if the overall structure of $P$ is compromised, the depthwise convolution is still able to capture the spatial features of the input. Interestingly, this is conceptually similar to the popular residual learning where $F_m^{in}$ is added to the module output, but ours uses $D \otimes F_m^{in} $ instead. By considering this residual term, we can finally formulate our low-rank pointwise residual module as:
\begin{equation}
 (P \times D )\otimes F^{\text{in}}_{m} \approx F^{P}_{m}+F^{{Res}}_{m} = (P^{(2)}_{mr}P^{(1)}_{rm}+I_{m}){D}_{m} \otimes F^{in}_{m},
\label{equ.8}
\end{equation}
where $I_m$ is an identity matrix. To further improve the performance, we may normalized the features of $F_m^{(P)}$ with L2 Normalization on the channel, and apply batch normalization on $D$.
With the factorization of the large matrix $P$, our LPR successfully reduces the parameters and computational costs comparing with other state-of-the-art modules.

\subsection{Ablation Study}

In this subsection we will first show the experiment of $r$ selection. Then an ablation study of our LPR module will be presented. At last we will validate our low-rank approach with an experiment on CIFAR-10.

\setlength{\tabcolsep}{1.2pt}
\begin{table}[!htp]
\setlength{\abovecaptionskip}{0.2 cm}
\setlength{\belowcaptionskip}{-0.5 cm}
\begin{center}

\scalebox{1}{
\begin{tabular}{l | c| c|c }
\toprule[1.2pt]
Rank of LPR& ImageNet Top-1&FLOPs  &Parameters\\
\hline
 $r = m/4$ & $70.8\%$ & $363$M & $3.0$M \\
$r = m/8$  &$70.6\%$ &$260$M & $2.3$M \\
$r = m/16$  & $68.1\%$ & $209$M & $2.0$M  \\
\hline
MobileNet  & $70.6\%$ & $574$M & $4.2$M  \\
\bottomrule[1.2pt]
\end{tabular}
}
\caption{Experiments to select the best rank $r$. $m$ means the number of the output channels. }\label{table:k-selection}
\end{center}
\end{table}

A set of experiments on ImageNet with MobileNet architecture has been performed to select the best rank $r$ during the low-rank decomposition which is shown in Equation \ref{equ.8}. The results are shown in Table \ref{table:k-selection}. From the table we can find that $r=m/8$ keep good performance while leads to significant reduction on the computational cost and parameters. With $r=m/8$ as the rank control parameter, the theoretical comparisons among the prevalent lightweight modules are shown in Table \ref{table:computation compare}. In this table, our LPR module has the least computational cost and parameters when the input and output are the same. Note that $4r <m-S_{\text{k}} ^{2}$ is the sufficient and necessary condition which can make LPR module have less computational cost and parameters than ShuffleNetv2 module. Thus, $r$ should be smaller than $m/4$. Note that $P^{(2)}_{mr}$ and $P^{(1)}_{rm}$ are learned to approximate the optimized matrices through training.

\setlength{\tabcolsep}{1.2pt}
\begin{table}[!htp]
\setlength{\abovecaptionskip}{0.2 cm}
\setlength{\belowcaptionskip}{-0.5 cm}
\begin{center}
\scalebox{0.9}{
\begin{tabular}{l | c| c }
\toprule[1.2pt]
Models & Accuracy & Paramters\\
\hline
SConv \cite{vgg}& $92.1\%$& $11$M\\

DSC Module\cite{mobilenets} & $91.8\%$ & $3.1$M \\

LPR Module(no Residual) & $88.5\%$&  $2.4$M\\

LPR Module(no L2Norm) & $90.6\%$ &  $2.4$M \\

LPR Module & $91.8\%$  & $2.4$M \\
\bottomrule[1.2pt]
\end{tabular}
}
\caption{Performance with different modules. Since the parameters are fixed during the training, the only updated modules are DSC and LPR. Therefore, the similar accuracy among different modules means the weight matrices are similar. }\label{table:CIFAR}
\end{center}
\end{table}

To validate the effectiveness of different parts in our LPR module, we train our LPRNet on the CIFAR-10 datasets after remove the L2 Normalization layer and residual part respectively. The comparison results are shown in the Table \ref{table:CIFAR}. From the table, it is clear that the completed LPR module has similar performance with the DSC module. However, its performance will drop after we remove the residual part. Besides, the performance will suffer a significant recession after we remove the L2 Normalization layer. Both parts will not increase the parameters of the model.

\begin{figure}[!t]
\setlength{\abovecaptionskip}{-0.1 cm}
\setlength{\belowcaptionskip}{-0.3cm}
\centering
\includegraphics[width=0.9\linewidth]{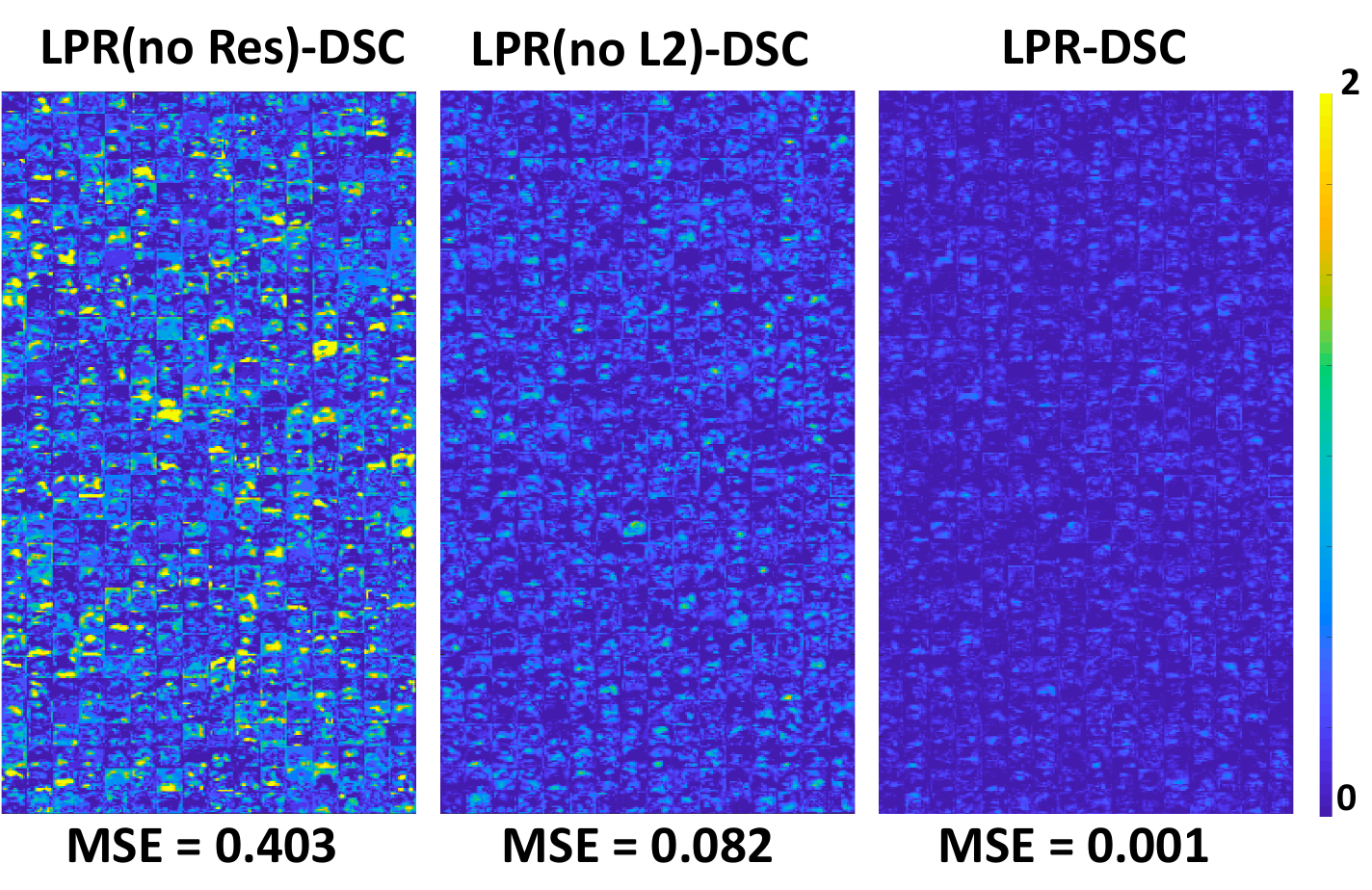}
\caption{The heatmap visualization of the difference among standard Convolution, DSC, and LPR. The similarity is represented by MSE. The lower MSE means the two feature matrices have higher similarity. Better viewed in color. }
\label{fig:lowrank_visualize}
\end{figure}

\setlength{\tabcolsep}{2pt}
\begin{table*}[!tp]
\begin{center}

\scalebox{0.85}{
\begin{tabular}{l | c| c| c| c| c| c| c}
\toprule[2pt]
Methods &Top-1& FLOPs & params&Methods &Top-1& FLOPs & params \\
\midrule[1.5pt]
ShuffleNetv2$_{\text{2018ECCV}}$\hspace{-0.8mm}$\times$\hspace{-0.8mm}1.0 \cite{shufflenetv2}& $ 69.3\% $ & 149M & 2.3M&
MobileNetv1$_{\text{2017 CoRR}}$\hspace{-0.8mm}$\times$\hspace{-0.8mm}1.0 \cite{mobilenets}& $ 70.6\% $ & 574M & 4.2M\\
\textbf{LPRNet}$_{\text{Shufflev2}}$\hspace{-0.2mm}$\times$\hspace{-0.2mm}1.0 & $ 69.1\% $ & 113M & 2.0M &
  \textbf{LPRNet}$_{\text{MobileNet}}$\hspace{-0.2mm}$\times$\hspace{-0.2mm}1.0 & $ 70.6\% $ & 260M & 2.3M\\

\midrule[2pt]
  ShuffleNetv1$_{\text{2018CVPR}}$\hspace{-0.8mm}$\times$\hspace{-0.8mm}0.25 \cite{Zhang_2018_shufflenet}& $ 38.5\% $ & 13M & 368K &
    MobileNetv1$_{\text{2017CoRR}}$\hspace{-0.8mm}$\times$\hspace{-0.8mm}0.25 \cite{mobilenets}& $ 50.6\% $ & 42M & 470K \\
ShuffleNetv2$_{\text{2018ECCV}}$ \hspace{-0.8mm}$\times$\hspace{-0.8mm}0.25 \cite{shufflenetv2}& $ 43.0\% $ & 14M & 587K &
  ShuffleNetv1$_{\text{2018CVPR}}$\hspace{-0.8mm}$\times$\hspace{-0.8mm}0.5 \cite{Zhang_2018_shufflenet}& $ 56.8\% $ & 41M & 718K\\
\textbf{LPRNet}$_{\text{MobileNet}}$\hspace{-0.2mm}$\times$\hspace{-0.2mm}0.25 & $ 50.1\% $ & 26M & 356K&

\textbf{LPRNet}$_{\text{MobileNet}}$\hspace{-0.2mm}$\times$\hspace{-0.2mm}0.5 & $ 63.2\% $ & 78M & 869K\\

\midrule[1pt]
IGCV3$_{\text{2018BMVC}}$\hspace{-0.8mm}$\times$\hspace{-0.8mm}$0.5$ \cite{igcv3}& $ 60.6\% $ & 111M & 2.0M &
ESPNetv2$_{\text{2019CVPR}}$\hspace{-0.8mm}$\times$\hspace{-0.8mm}1.0 \cite{espnetv2}& $ 64.6\% $ & 98M & 1.7M\\
MobileNetv1$_{\text{2017CoRR}}$\hspace{-0.8mm}$\times$\hspace{-0.8mm}0.5 \cite{mobilenets}& $ 63.7\% $ & 152M & 1.3M&
ESPNetv2$_{\text{2019CVPR}}$\hspace{-0.8mm}$\times$\hspace{-0.8mm}1.25 \cite{espnetv2}& $ 66.8\% $ & 138M & 1.9M\\
 MobileNetv2$_{\text{2018CVPR}}$ \hspace{-0.8mm}$\times$\hspace{-0.8mm}0.5 \cite{Mobilenetv2}& $ 64.3\% $ & 100M & 1.9M&
 \textbf{LPRNet}$_{\text{Shufflev2}}$\hspace{-0.2mm}$\times$\hspace{-0.2mm}1.0 & $ 69.1\% $ & 113M & 2.0M\\
 \midrule[1pt]
 ShuffleNetv1$_{\text{2018CVPR}}$\hspace{-0.8mm}$\times$\hspace{-0.8mm}1.0 \cite{Zhang_2018_shufflenet}& $ 67.4\% $ & 140M & 2.2M&
  ShuffleNetv2$_{\text{2018ECCV}}$\hspace{-0.8mm}$\times$\hspace{-0.8mm}1.0 \cite{shufflenetv2}& $ 69.3\% $ & 149M & 2.3M\\
 ESPNetv2$_{\text{2019CVPR}}$\hspace{-0.8mm}$\times$\hspace{-0.8mm}1.5 \cite{espnetv2}& $ 67.9\% $ & 185M & 2.3M & 
 CondenseNet$_{\text{2018CVPR}}$ \cite{condensenet}& $ 70.3\% $ & 291M & 2.9M\\
IGCV3$_{\text{2018BMVC}}$\hspace{-0.8mm}$\times$\hspace{-0.8mm}0.75 \cite{igcv3}& $ 69.1\% $ & 210M & 2.6M& 
 \textbf{LPRNet}$_{\text{MobileNet}}$\hspace{-0.2mm}$\times$\hspace{-0.2mm}1.0 & $ 70.6\% $ & 260M & 2.3M\\ 
 \midrule[1pt]
MobileNetv1$_{\text{2017CoRR}}$\hspace{-0.8mm}$\times$\hspace{-0.8mm}0.75 \cite{mobilenets}& $ 68.4\% $ & 325M & 3.6M &
 IGCV3$_{\text{2018BMVC}}$\hspace{-0.8mm}$\times$\hspace{-0.8mm}1.0 \cite{igcv3}& $ 71.7\% $ & 340M & 3.4M \\

ESPNetv2$_{\text{2019CVPR}}$\hspace{-0.8mm}$\times$\hspace{-0.8mm}2.0 \cite{espnetv2}& $ 71.0\% $ & 306M & 3.5M &
 MobileNetv2$_{\text{2018CVPR}}$\hspace{-0.8mm}$\times$\hspace{-0.8mm}1.0 \cite{Mobilenetv2}& $ 72.0\% $ & 301M & 3.5M \\

ShuffleNetv1$_{\text{2018CVPR}}$\hspace{-0.8mm}$\times$\hspace{-0.8mm}1.5 \cite{Zhang_2018_shufflenet}& $ 71.5\% $ & 292M & 3.4M &
\textbf{LPRNet}$_{\text{MobileNet}}$\hspace{-0.2mm}$\times$\hspace{-0.2mm}1.25 & $ 72.8\% $ & 389M & 3.4M \\
 \midrule[1pt]

MobileNetv1$_{\text{2017 CoRR}}$\hspace{-0.8mm}$\times$\hspace{-0.8mm}1.0 \cite{mobilenets}& $ 70.6\% $ & 574M & 4.2M &
ShuffleNetv2$_{\text{2018ECCV}}$\hspace{-0.8mm}$\times$\hspace{-0.8mm}2.0 \cite{shufflenetv2}& $ 74.1\% $ & 595M & 7.6M \\
ShuffleNetv1$_{\text{2018CVPR}}$\hspace{-0.8mm}$\times$\hspace{-0.8mm}2.0 \cite{Zhang_2018_shufflenet}& $ 73.4\% $ & 524M & 5.4M &
 MobileNetv2$_{\text{2018CVPR}}$\hspace{-0.8mm}$\times$\hspace{-0.8mm}1.4 \cite{Mobilenetv2}& $ 74.4\% $ & 585M & 6.9M \\  
\textbf{LPRNet}$_{\text{Shufflev2}}$\hspace{-0.2mm}$\times$\hspace{-0.2mm}2.0 & $ 73.8\% $ & 437M & 6.2M &
 \textbf{LPRNet}$_{\text{MobileNet}}$\hspace{-0.2mm}$\times$\hspace{-0.2mm}1.5 & $ 74.6\% $ & 544M & 4.5M \\

\bottomrule[2pt]
\end{tabular}
}
\caption{Performance of image classification task on ImageNet. The first region shows direct comparisons between the LPRNet and its underlying structures; the rest regions are divided based on the size of parameters (from smallest to largest).}\label{table:imagenettask}
\end{center}
\end{table*}

We also design an experiment to verify the ability of the Low-Rank approach of our LPR module. In our experiment, a network using standard convolution is trained on CIFAR-10. Then we replace a layer whose dimension is $512\times 512\times14\times14$ with DSC module, LPR without Residual, LPR without L2 Normalization, and our LPR module respectively. The network is trained while all the other parameters are fixed. Training process is stopped when the model has similar Top-1 validation accuracy with the original network. The similarities among output features are represented by Mean Squire Error (MSE) and visualized by the heatmaps. The results are shown in Figure \ref{fig:lowrank_visualize}. As shown in the figure, the MSE between LPR and DSC is only $0.001$, which means the output features from LPR and DSC have little difference. Since the other parameters of the network are fixed, similar output features mean the weights of the two modules are similar, which supports that the weight matrix of our low-rank decomposition structure can approach the matrix of Depthwise Separable Convolution. We can also find from the figure that the MSE increases when we remove the residual part and the L2 Normalization. Furthermore, the MSE increases more than $400$ times when the residual part is removed from our LPR module, which indicates that the learned weights is hard to approach the weights of the DSC module.



\subsection{Implementations}
\begin{figure}[!t]
\setlength{\abovecaptionskip}{-0.1 cm}
\setlength{\belowcaptionskip}{-0.3cm}
\centering
\includegraphics[width=1\linewidth]{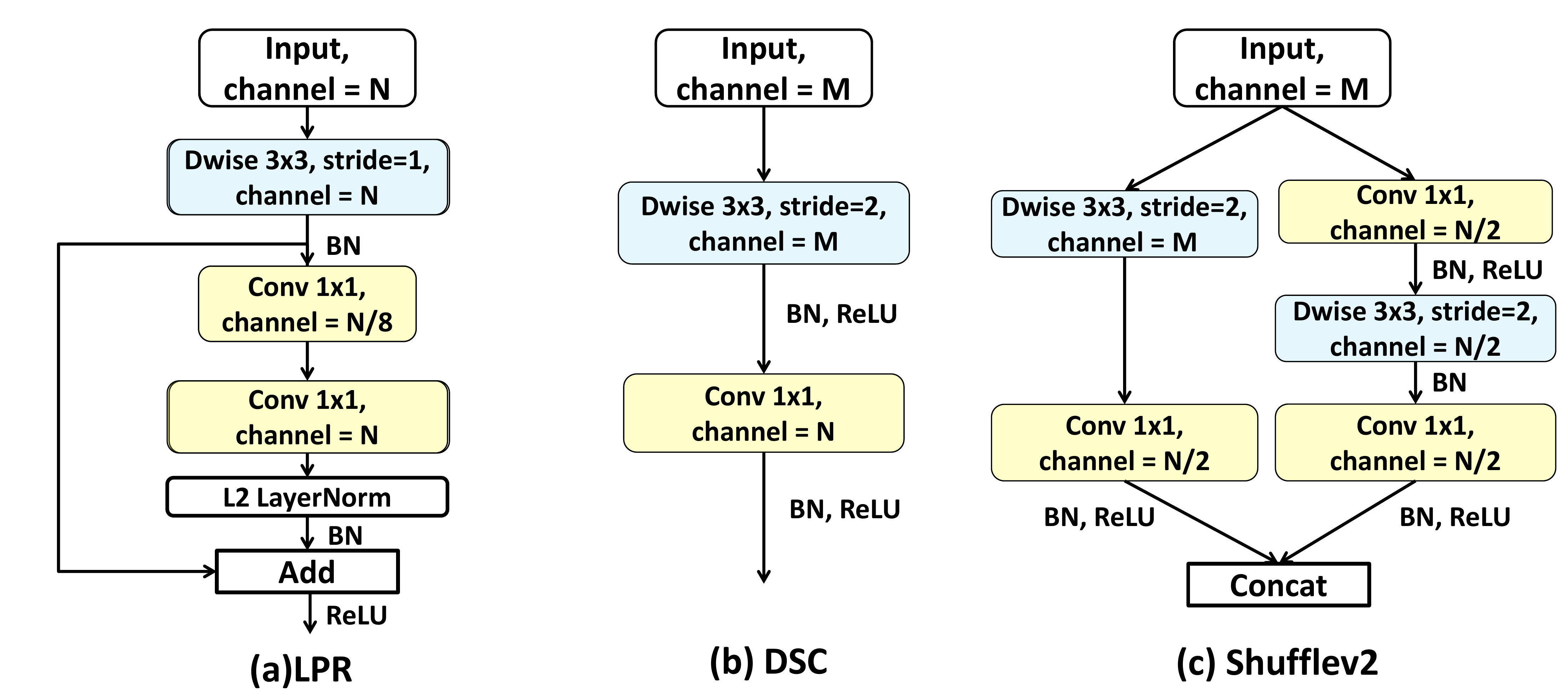}
\caption{(a): Our LPR module. (b): Down-sample module using Depthwise Separable Convolution. (c): Down-sample module using ShuffleNetv2 module. Note that DSC is also used when the input and output of the module are not in the same dimension.}
\label{fig:basicmodule}
\end{figure}

In this subsection, we embody the LPRNet based on our LPR module and the deep learning structure of MobileNetv1 and ShuffleNetv2, respectively.
The reason for choosing MobileNet \cite{mobilenets} and ShuffleNetv2 \cite{shufflenetv2} is that most modules of these two networks have the same input dimension and output dimension, which is the condition to utilize our LPR module. The details of the module we used in our LPRNet are shown in Figure \ref{fig:basicmodule}. 
Since the input of the LPR module requires identical input-output dimension, the down sample modules of the MobileNetv1 and ShuffleNetv2 are reserved in our LPRNet. The rest modules are replaced by our LPR module.

\begin{figure*}[t]
\setlength{\abovecaptionskip}{0.cm}
\setlength{\belowcaptionskip}{-0.3cm}
\begin{center}
\includegraphics[width=0.9\linewidth]{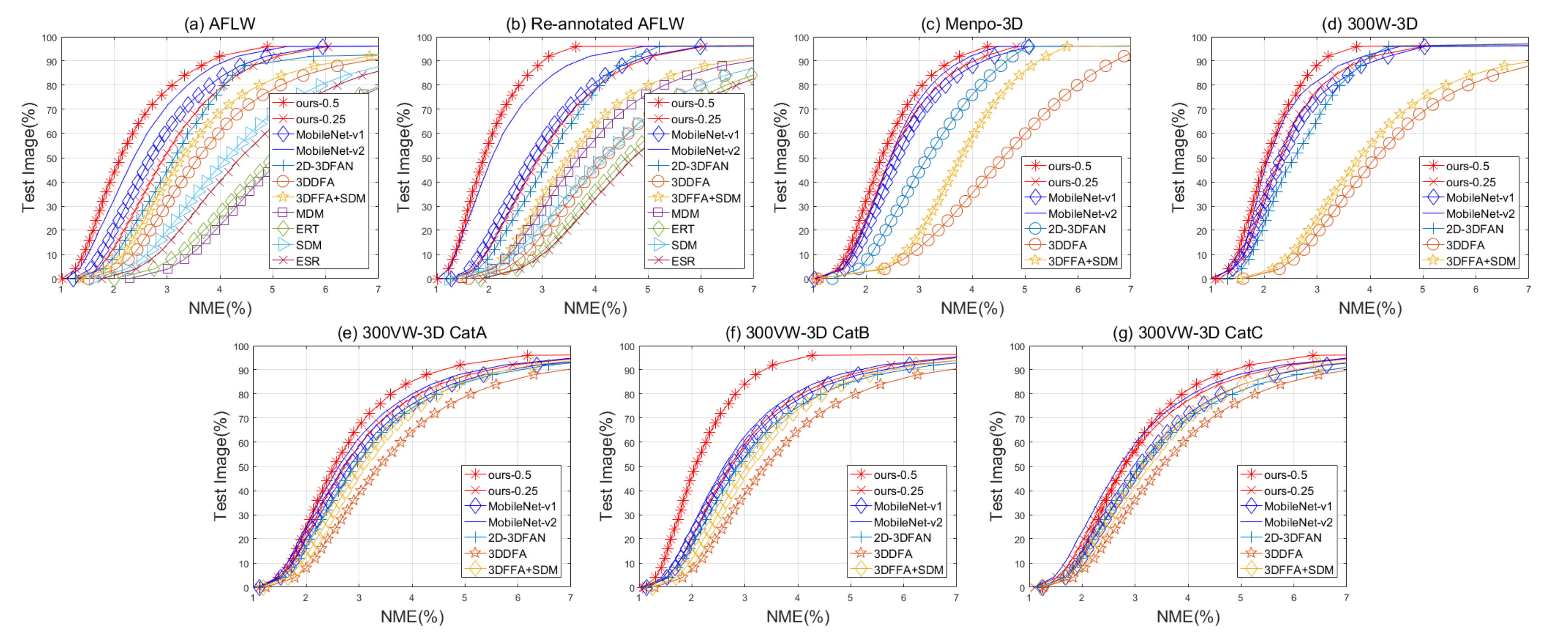}
\end{center}
\caption{The CED curves on different test datasets.We test the 3D methods on whole datasets of AFLW2000-3D, re-annotated AFLW2000-3D, Menpo-3D, 300W-3D, and 300VW-3D. The top-left curve means the method has the best performance, i.e., most cases below given NME. }
\label{fig:experimentsresults}
\end{figure*}

\section{Experiment}
In this section, we conduct experiments on image classification and large poses Face Alignment tasks. In each subsection, we present datasets, comparison methods, parameter settings, evaluation metrics, and show comparison results.

\subsection{Image Classification}

\textbf{Dataset:} To make an fair comparison, the dataset we use is ImageNet 2012 classification dataset \cite{deng2009imagenet,russakovsky2015imagenet}. There are $128,1167$ images and $1,000$ classes in training dataset. The images in training dataset are resized to $480 \times 480$ and are randomly cropped. The images in validation dataset are resized to $256 \times 256$ and are center cropped. Some augmentations such as random flip, random scale, and random illumination are implemented on the training dataset. All the results are tested on the validation dataset.

\textbf{Comparison Methods:} In this secion, we compare with lightweight architectures on ImageNet classification including MobileNetv1 \cite{mobilenets}, ShuffleNetv1 \cite{Zhang_2018_shufflenet}, MobileNetv2 \cite{Mobilenetv2}, ShuffleNetv2 \cite{shufflenetv2}, MnasNet \cite{tan2018mnasnet}, CondenseNet  \cite{condensenet}, IGCV3 \cite{igcv3}, and ESPNetv2 \cite{espnetv2}.

\textbf{Parameter Settings:} Our learning model is built by Mxnet framework \cite{chen2015mxnet}. The optimizer is the large batch SGD \cite{lbsgd} starting with the learning rate $0.5$. The learning rate is decayed following cosine function. The total epoch number is set to 210 for LPR$_{\text{MobileNet}}$, and 400 for LPR$_{\text{Shufflev2}}$. The batch size is set to 256 for LPR$_{\text{MobileNet}}$ and 400 for LPR$_{\text{Shufflev2}}$. After training, the model is tuned on the same training dataset without data augmentation.

\textbf{Evaluation Metrics:} In this paper, we evaluate the performance using Top-1 accuracy. Like other works \cite{Mobilenetv2,shufflenetv2,tan2018mnasnet,netadapt}, the computational cost is evaluated using calculated FLOPs number and the parameters are evaluated by the calculated number.

\textbf{Comparison Results:} The comparison results are shown in the Table \ref{table:imagenettask}. The table is divided into six regions. The first region is the direct comparison between the LPRNet and its underlying architecture from MobileNetv2 and ShuffleNetv2. The rest regions are divided based on parameters. Compared with the same architecture, our LPR$_{\text{MobileNet}}$ can retain the same performance while reduce the computational cost and parameters significantly. Though the LPR$_{\text{Shufflev2}}$ has $0.9\%$ a lower accuracy than ShuffleNetv2, it is only $83\%$ of the original size and uses $75\%$ fewer Flops than ShuffleNetv2. Compared with other methods, the LPRNet also achieves the best performance with approximately the same complexity. When the parameters are reduced to the \textit{K} level, our LPRNet has over $63\%$ Top-1 accuracy while the accuracy of all other methods is below $57\%$. Even the parameters are reduced to the least, we can still achieve $50\%$ accuracy. The MobileNetv2$\times 0.25$ and ShuffleNetv2$\times0.5$ are not shown in the K level since their parameters are largerh than $1$M.

\setlength{\tabcolsep}{2pt}
\begin{table*}[t!]\footnotesize
\begin{center}
\setlength{\abovecaptionskip}{0.1 cm}
\setlength{\belowcaptionskip}{-0.5 cm}
\scalebox{0.9}{
\begin{tabular}{l |c|c|c|c|c|c|c}
\toprule[1.2pt]
 \multicolumn{5}{c|}{Normalized Mean Error on AFLW2000-3D} & \multicolumn{2}{c|}{Speed (FPS)}& Memory \\
 \hline
 Method Name&$[0^\circ 30^\circ]$&$[30^\circ 60^\circ]$&$[60^\circ 90^\circ]$&Mean&GPU &CPU &params (Bytes)\\
 \hline
  \textbf{LPRNet$_{\text{MobileNet}}$ $\times 0.25$}&\textbf{2.58}&\textbf{3.33}&\textbf{5.81} & \textbf{3.89}&\textbf{800}&\textbf{95}&\textbf{619K}\\
  \textbf{LPRNet$_{\text{MobileNet}}$ $\times 0.5$}&\textbf{2.41}&\textbf{3.15}&\textbf{5.52} & \textbf{3.69}&\textbf{630}&\textbf{52}&\textbf{1.6M}\\

 ShuffleNetv2$_{\text{2018ECCV}}$ \hspace{-0.8mm}$\times$\hspace{-0.8mm}0.5 \cite{shufflenetv2}&2.54&3.32&5.51  & 3.79 &$600$&51&3.3M\\
 ShuffleNetv1$_{\text{2018CVPR}}$\hspace{-0.8mm}$\times$\hspace{-0.8mm}0.5 \cite{Zhang_2018_shufflenet}&3.03	&4.01&6.07	  & 4.37 &$900$&104&2.1M\\
 MobileNetv2$_{\text{2018CVPR}}$ \hspace{-0.8mm}$\times$\hspace{-0.8mm}0.5 \cite{Mobilenetv2}&2.51&3.15&5.53 & 3.73&600&40&2.3M\\
MobileNetv1$_{\text{2017CoRR}}$\hspace{-0.8mm}$\times$\hspace{-0.8mm}0.5 \cite{mobilenets}&2.52&3.21&5.76 & 3.85&600&42&2.5M\\

PCD-CNN$_{\text{2018CVPR}}$ \cite{kumar2018disentangling} &2.93&4.14&4.71  & 3.92 &20&-&-\\
Hyperface$_{\text{2017TPAMI}}$ \cite{hyperface} &3.93&4.14&6.71  & 4.26&-&-&119.7M\\
3DSTN$_{\text{2017ICCV}}$ \cite{3DSTN} &3.15&4.33&5.98  & 4.49&52&-&-\\
 3DFAN$_{\text{2017ICCV}}$ \cite{bulat2017far}&2.75&3.76&5.72  &4.07&6&$<1$&183M\\
 3DDFA$_{\text{2016CVPR}}$ \cite{zhu2016face} &3.78&4.54&7.93  & 5.42&43&13&111M\\
3DDFA+SDM$_{\text{2016CVPR}}$ \cite{zhu2016face}&3.43&4.24&7.17  & 4.94&31&9&121M\\
 MDM$_{\text{2016CVPR}}$ \cite{trigeorgis2016mnemonic} &3.67&5.94&10.76 &6.45 &5&$<1$&307M\\
 ERT$_{\text{2014CVPR}}$ \cite{kazemi2014one}&5.40&7.12&16.01    &10.55 &-&300&95M\\
 ESR$_{\text{2014IJCV}}$ \cite{esr}&4.60&6.70&12.67 &7.99 &-&83&248M\\
  SDM$_{\text{2013CVPR}}$ \cite{xiong2013supervised}&3.67&4.94&9.76   &6.12 &-&80&10M\\
\bottomrule[1.2pt]
\end{tabular}
}
\caption{Comparisons with state-of-the-art methods on AFLW2000-3D dataset.  }\label{table:comparison}%
\end{center}

\end{table*}

\subsection{Large Poses Face Alignment}

\textbf{Datasets:} In our face alignment experiments, all the baselines use 68-point landmarks to conduct fair comparisons. We evaluate all the baselines with only x-y coordinates for fair comparisons, since some datasets \cite{bulat2017far} used only have 2D coordinates projected from 3D landmarks. Training datasets are 300W-LP \cite{zhu2016face}, while testing datasets are AFLW2000-3D \cite{zhu2016face}, Re-annotated AFLW2000-3D \cite{bulat2017far},LS3D-W \cite{bulat2017far} which has 5 sub-dataset Menpo-3D (8,955 images), 300W-3D (600 images), and 300VW-3D (A, B, and C). 

\textbf{Comparison Methods:} We conduct comprehensive evaluations with the state-of-the-art methods. In this section, a comparison is made with state-of-the-art deep methods including PCD-CNN \cite{kumar2018disentangling}, 3DFAN \cite{bulat2017far}, Hyperface \cite{hyperface}, 3DSTN \cite{3DSTN}, 3DDFA  \cite{zhu2016face}, and MDM \cite{trigeorgis2016mnemonic}. Among these baselines, 
the results of 3DSTN and PCD-CNN are cited from the original papers. We also compare the accuracy and speed on CPU, with some those methods only running on CPU, including SDM \cite{xiong2013supervised}, ERT \cite{kazemi2014one}, and ESR \cite{esr}. To make a fair comparison, the lightweight models MobileNet \cite{mobilenets,Mobilenetv2} and ShuffleNet \cite{Zhang_2018_shufflenet,shufflenetv2} are implemented for face alignment and are trained on the same datasets. All these models are using half channels for fast training and testing.

\textbf{Parameter Settings:} Our structure is built by Mxnet framework \cite{chen2015mxnet} and uses $L_{2}$ loss specified for regression task. We use Adam stochastic optimization \cite{kingma2014adam} with default hyper-parameters to learn the weights. The initial learning rate is set to 0.0005, and the initialized weights are generated with Xavier initialization. The epoch is set to 60, and the batch size is set to 100. We set the learning rate to $4e^{-4}$  at first 15 epoch and then decay the learning rate to $2e^{-4}$ when the number of channels is multiplied by $0.5$.

\textbf{Evaluation Metrics:} We use ground-truth landmarks to generate bounding boxes. ``Normalized Mean Error (NME)'' is an important metric for face alignment evaluation, which is defined as $\text{NME}=\frac{1}{N}\sum_{i=1}^{N}\frac{\left \|\hat{X}_{i}-X_{i}^*\right \|_{2}}{d}$ where the $\hat{X}$ and $X^*$ is predicted and ground truth landmarks, respectively, and $N$ is the number of the landmarks. $d$ can be computed using $d=\sqrt{w_{\text{bbox}}\times h_{\text{bbox}}}$, in which $w_{\text{bbox}}$ and $h_{\text{bbox}}$ mean the width and height of the bounding box, respectively. 
The speed of all methods is evaluated on Intel-i7 CPU without Openmpi. We use Frames Per Second (FPS) to evaluate the speed. The storage size in this paper is calculated from the compressed model.

\begin{figure}[t]
\setlength{\abovecaptionskip}{0.cm}

\centering
\setlength{\abovecaptionskip}{0.1 cm}
\setlength{\belowcaptionskip}{-0.3 cm}
\includegraphics[width=0.9\linewidth]{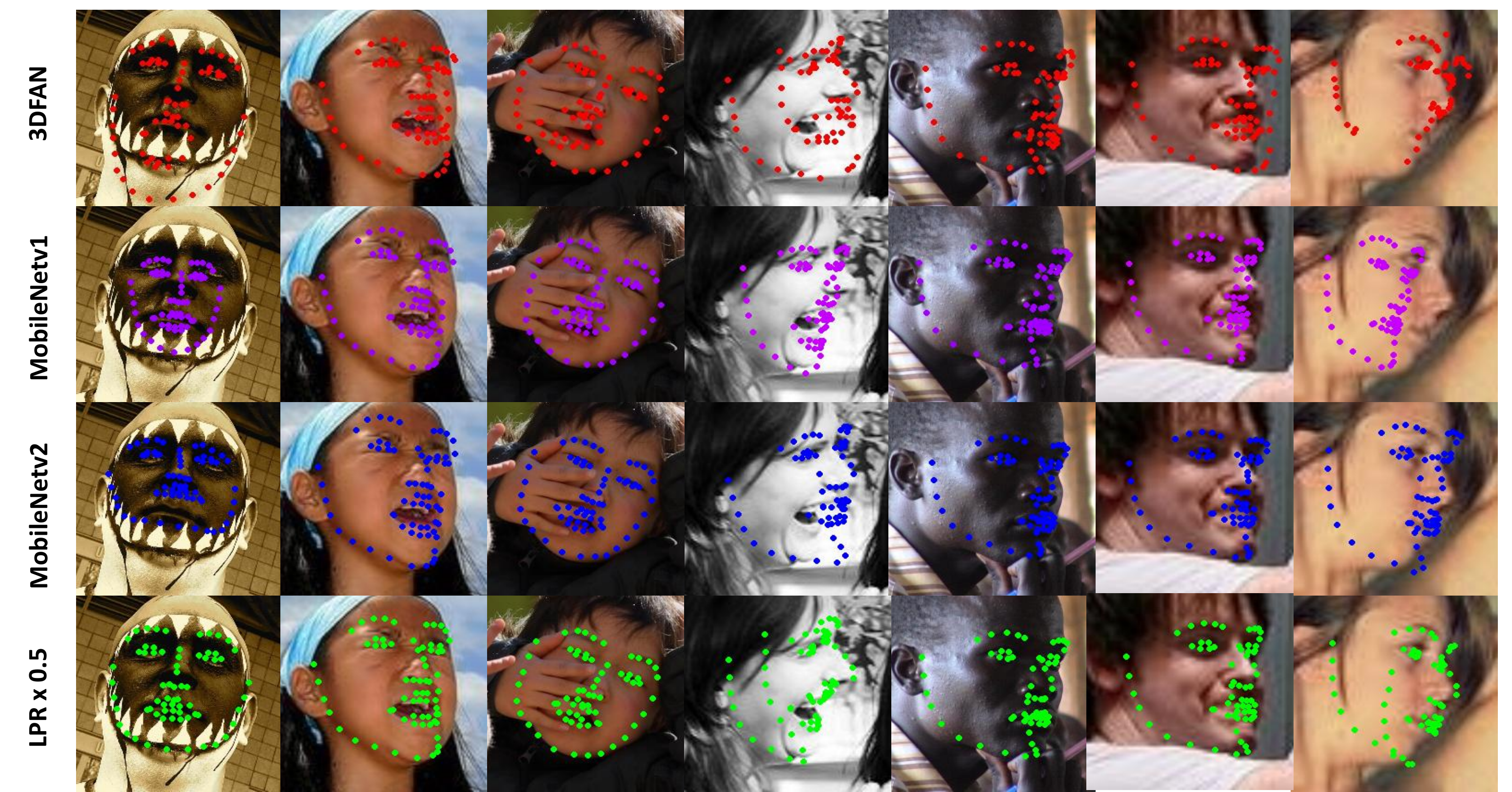}
\caption{Some visualized results. }
\label{fig:visualization}
\end{figure}

\textbf{Comparison Results:} To compare the performance of the different range of angles, we divide the testing dataset into three parts by the range of the angles of the faces \cite{zhu2016face}. The curve of the cumulative errors distribution (CED) of the whole dataset is shown in Figure \ref{fig:experimentsresults}.
The visualization comparison results are shown in Figure \ref{fig:visualization}. From the Table \ref{table:comparison},  we can observe that the NME of our LPRNet $\times 0.5$ is $5\%$ lower than the current state-of-the-art PCD-CNN. The NME of our LPRNet $\times 0.25$ achieves similar performance as PCD-CNN and MobileNetv1. Comparison with other lightweight model is also shown in the table. Comparing with MobileNetv1 and MobileNetv2, our LPRNet $\times 0.25$ has similar NME but with $\times 1.8$ speed on CPU and $73\%$ compression ratio.

\emph{Time Complexity:} Compared with those traditional deep learning methods \cite{hyperface,3DSTN,zhu2016face,zhu2016face,bulat2017far,hinton2015distilling,trigeorgis2016mnemonic}, our LPRNet has much better speed on both one core CPU and GPU. In the table, we notice that the SDM \cite{xiong2013supervised}, ERT \cite{kazemi2014one} and ESR \cite{esr} have very impressive speed on CPU. The reason is all of these methods use hand-craft features, which are easy to compute by computers but have limited ability for representation. comparing with the light weight models \cite{mobilenets,Mobilenetv2,Zhang_2018_shufflenet,shufflenetv2}, our LPRNet achieves the second best speed on CPU when the parameters are down-scaled by $0.25$, but get much better performance than the fastest model \cite{Zhang_2018_shufflenet}.

\emph{Space Complexity:} For the applications on mobile devices, the memory size of the model should be smaller enough. From the Table \ref{table:comparison}, it can be observed that our LPRNet is $\times 120$ smaller than the smallest model in baseline deep learning methods except for MobileNet. Besides, it is $\times 3.4$ smaller than the smallest model ShuffleNetv1 with much lower NME.

\section{Conclusion}

In this paper, we proposed a novel lightweight deep learning module named LPR to further reduce the network parameters through low-rank matrix decomposition and residual learning. By applying our LPR module to MobileNet and ShuffleNetv2, we managed to reduce the size of existing lightweight models. We surprisingly found that on image classification and face alignment tasks, compared to many state-of-the-art deep learning models, our LPRNet had much lower parameters and computational cost, but kept very competitive or even better performance. More importantly, this casts a light on deep models compression through low-rank matrix decomposition, and enables many powerful deep models to be deployed in end devices. We plan to release our models (in Mxnet) upon the publication of this work.


\medskip

\newpage
\fontsize{7.4pt}{8.4pt} \selectfont
\bibliography{egbib}
\bibliographystyle{aaai}

\end{document}